\documentclass[letterpaper, 10 pt, conference]{ieeeconf}  

\IEEEoverridecommandlockouts                              

\usepackage{cite}
\usepackage{graphicx}
\usepackage{float}
\usepackage{amsmath}
\usepackage{amssymb}
\usepackage{tabularx}
\usepackage{booktabs}
\usepackage{multirow}
\usepackage{amsfonts}
\usepackage{pdfpages}
\usepackage{xcolor}

\title{\LARGE \bf
ThermalLoc: A Vision Transformer-Based Approach for Robust Thermal Camera Relocalization in Large-Scale Environments
}

\author{Yu Liu$^{1}$, Yangtao Meng$^{1}$, Xianfei Pan$^{1}$, Jie Jiang$^{2}$ and Changhao Chen*$^{3}$
\thanks{$^{1}$ Yu Liu, Yangtao Meng and Xianfei Pan are with the College of Intelligence Science and Technology, National University of Defense Technology, Changsha, 410073, China}%
\thanks{$^{2}$ Jie Jiang is with the China Academy of Launch Vehicle Technology, Beijing, 100076, China}%
\thanks{$^{3}$ Changhao Chen is with PEAK-Lab, The Hong Kong University of Science and Technology (Guangzhou), Guangzhou, 511453, China}%
\thanks{*Changhao Chen is the corresponding author (email: changhaochen@hkust-gz.edu.cn)}%
}

\begin{document}

\maketitle
\thispagestyle{empty}
\pagestyle{empty}

\begin{abstract}
Thermal cameras capture environmental data through heat emission, a fundamentally different mechanism compared to visible light cameras, which rely on pinhole imaging. As a result, traditional visual relocalization methods designed for visible light images are not directly applicable to thermal images. Despite significant advancements in deep learning for camera relocalization, approaches specifically tailored for thermal camera-based relocalization remain underexplored. To address this gap, we introduce ThermalLoc, a novel end-to-end deep learning method for thermal image relocalization. ThermalLoc effectively extracts both local and global features from thermal images by integrating EfficientNet with Transformers, and performs absolute pose regression using two MLP networks. We evaluated ThermalLoc on both the publicly available thermal-odometry dataset and our own dataset. The results demonstrate that ThermalLoc outperforms existing representative methods employed for thermal camera relocalization, including AtLoc, MapNet, PoseNet, and RobustLoc, achieving superior accuracy and robustness.
\end{abstract}

\section{INTRODUCTION}
Camera relocalization or absolute pose estimation is crucial in autonomous driving \cite{autonomous-navigation}, robot navigation, as well as virtual and augmented reality \cite{AR}. Traditional camera relocalization methods \cite{orbslam3}, \cite{orbslam}, \cite{lsdslam} build a feature map and rely on feature extraction between images and the map. However, these methods often struggle under challenging conditions such as illumination variations, texture-less surfaces, and dynamic lighting conditions \cite{dynamic-lighting}. Radar-based methods \cite{loam}, \cite{pointloc}, \cite{hypliloc} also degrade in environments with airborne particulates. 
Thermal imaging addresses many of these limitations by detecting the intensity distribution of thermal radiation emitted from object surfaces. It offers two key advantages: it is unaffected by ambient lighting conditions, as it directly measures thermal signatures, and it has a superior ability to penetrate atmospheric obscurants (e.g., fog, smoke, and dust). As a result, thermal camera relocalization provides advantages such as all-day adaptability, stable operation in low-light or smoky environments, and robustness to dynamic lighting and certain weather conditions, particularly during diurnal shifts or when vehicle lights interfere.


However, developing thermal camera-based localization systems is challenging due to the lower resolution (typically converted from 14-bit to 8-bit), reduced texture, and contrast of thermal images, making traditional feature extraction less effective. In large-scale scenarios, thermal radiation attenuation degrades the signal-to-noise ratio (SNR), impairing feature extraction for distant objects. Periodic non-uniformity correction (NUC), required every 500 ms, introduces viewpoint shifts and increases odometry track loss. Additionally, global feature matching becomes computationally expensive as scene scale and feature density grow. Thermal imaging also suffers from spatiotemporal inconsistencies due to environmental interference, with materials exhibiting varying thermal signatures based on time of day and viewing angle.
\begin{figure}[t]
 \centering
 \includegraphics[width=0.5\textwidth]{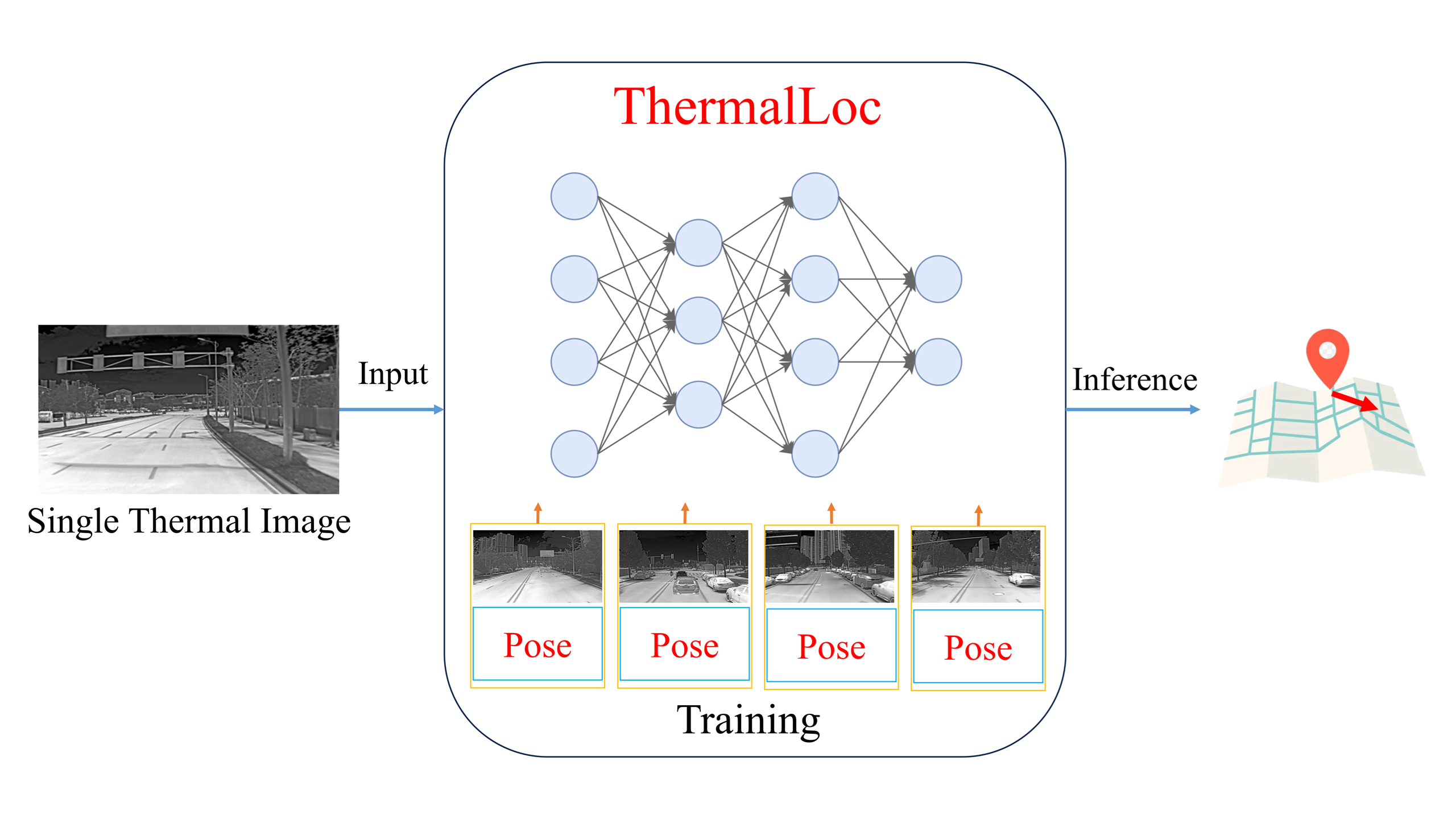}
 \caption{In our ThermalLoc model, we trained an efficient Transformer-based network on a thermal image dataset, requiring only a single thermal image input to accurately infer the 6-DoF pose.}
 \label{Concept map}
\end{figure}

In addition to traditional feature-based localization methods, end-to-end approaches for direct 6-DoF pose recovery, have been explored. PoseNet \cite{posenet} was an early APR model that uses a modified GoogleNet \cite{googlenet} to regress camera poses. Subsequent enhancements have included Bayesian methods \cite{BYS-based-posenet}, LSTM \cite{LSTM-based-posenet}, and other techniques. MapNet \cite{mapnet} integrates spatial geometric constraints between frames to enhance motion consistency in pose regression, while MS-Transformer \cite{MS-Transformer} applies Transformer-based methods to learn multi-scene APR. 
Currently, APR research has predominantly focused on RGB or lidar inputs, with thermal-based APR remaining relatively unexplored. While some literature addresses indoor dark environments using thermal images\cite{darkloc+}, there is a notable lack of research on thermal-based APR for large-scale scenes.

\begin{figure*}[t]
 \centering
 \includegraphics[width=0.9\textwidth,height=8.5cm]{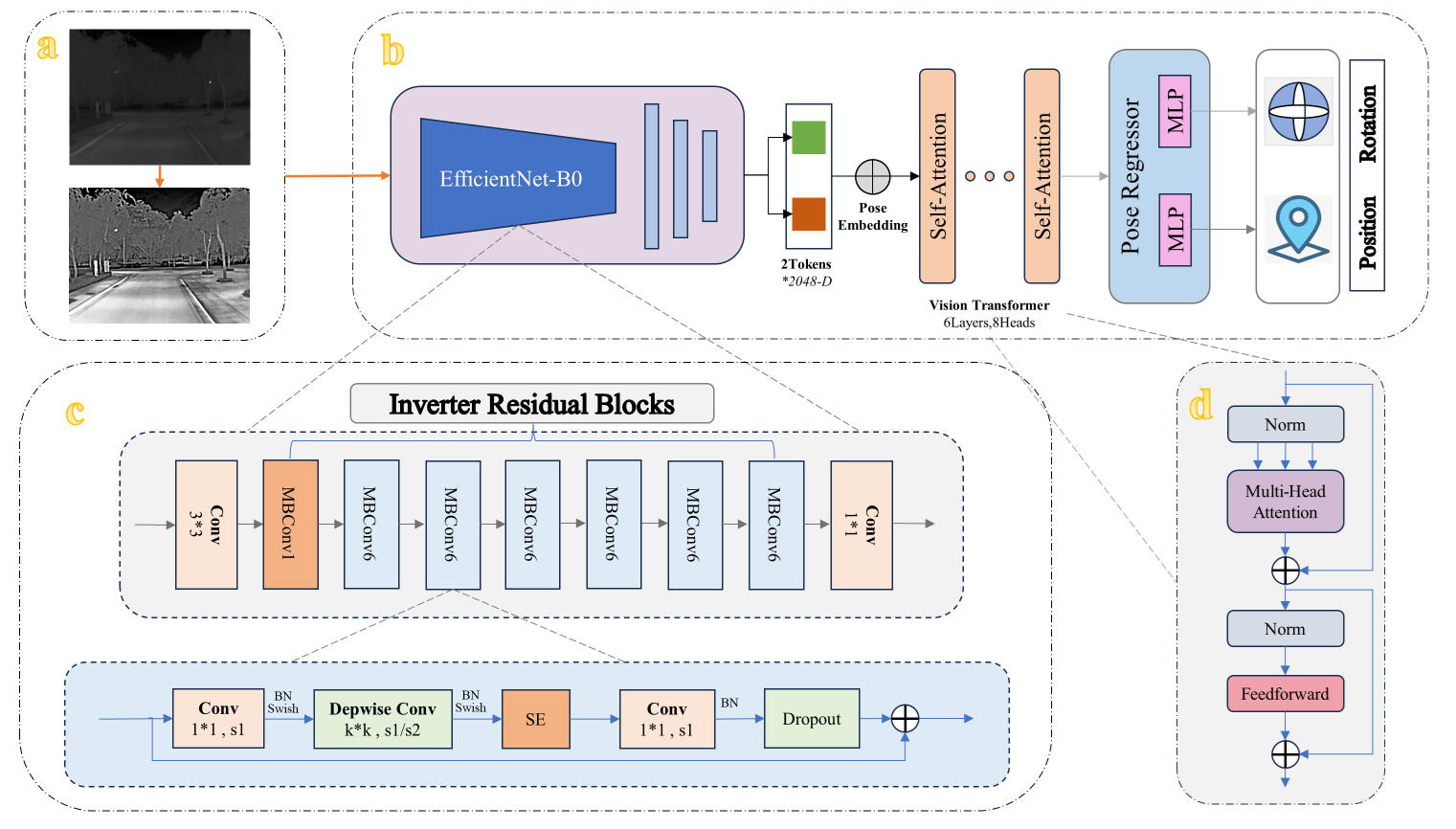}
 \caption{(a) Thermal image preprocessing: the processed thermal image (below) is compared with the original thermal image (above). The processed image is then input into (b) the ThermalLoc model framework. This framework consists of an EfficientNet-B0 module for extracting local features, a 6-layer Transformer for capturing global feature associations, and an MLP-based regressor to predict the 6-DoF camera pose. (c) The architecture of the EfficientNet-B0 module (above) and the MBConv block (below) used within EfficientNet. (d) The architecture of a self-attention block in Transformer layer.}
 \label{fig:framework}
\end{figure*}

To address this gap, we propose \textbf{ThermalLoc}, a novel thermal-based APR framework that employs EfficientNet \cite{efficientnet} and Vision Transformer \cite{vit} for efficient 6-DoF camera relocalization. EfficientNet enhances the local feature extraction capabilities of ThermalLoc, improving robustness across diverse scenes. The Vision Transformer further captures global feature associations through its self-attention mechanism, making our model well-suited for large-scale scenarios. We introduce a connectivity method that processes the EfficientNet feature map before integrating it with the Transformer, optimizing the thermal image extraction process. Additionally, we optimize thermal images to improve relocalization accuracy. We adjust brightness and contrast through linear transformation and enhance image detail with a Gaussian low-pass filter. Extensive real-world experiments and a challenging thermal image dataset were used to evaluate ThermalLoc. Our results demonstrate that ThermalLoc significantly outperforms the representative end-to-end relocalization method, such as AtLoc, and RobustLoc in terms of positioning accuracy.

Our contributions are summarized as follows:
\begin{itemize}
\item We propose ThermalLoc, the first end-to-end learning model that performs camera relocalization in city-scale scenarios using only single thermal images.
\item  To enhance feature extraction in thermal images, we introduce a combination of EfficientNet and Transformer with an efficient connectivity method that simultaneously captures both global and local features. This approach proves particularly effective for thermal camera relocalization.
\item We developed a challenging thermal image dataset captured from a vehicle in city-scale, featuring dynamic environments, complex lighting conditions, and pedestrian interference. Experimental results demonstrate that our proposed method exhibits robustness and accuracy, demonstrating the state-of-the-art performance in thermal camera relocalization.
\end{itemize}

\begin{figure*}[t]
 \centering
 \includegraphics[width=1\textwidth]{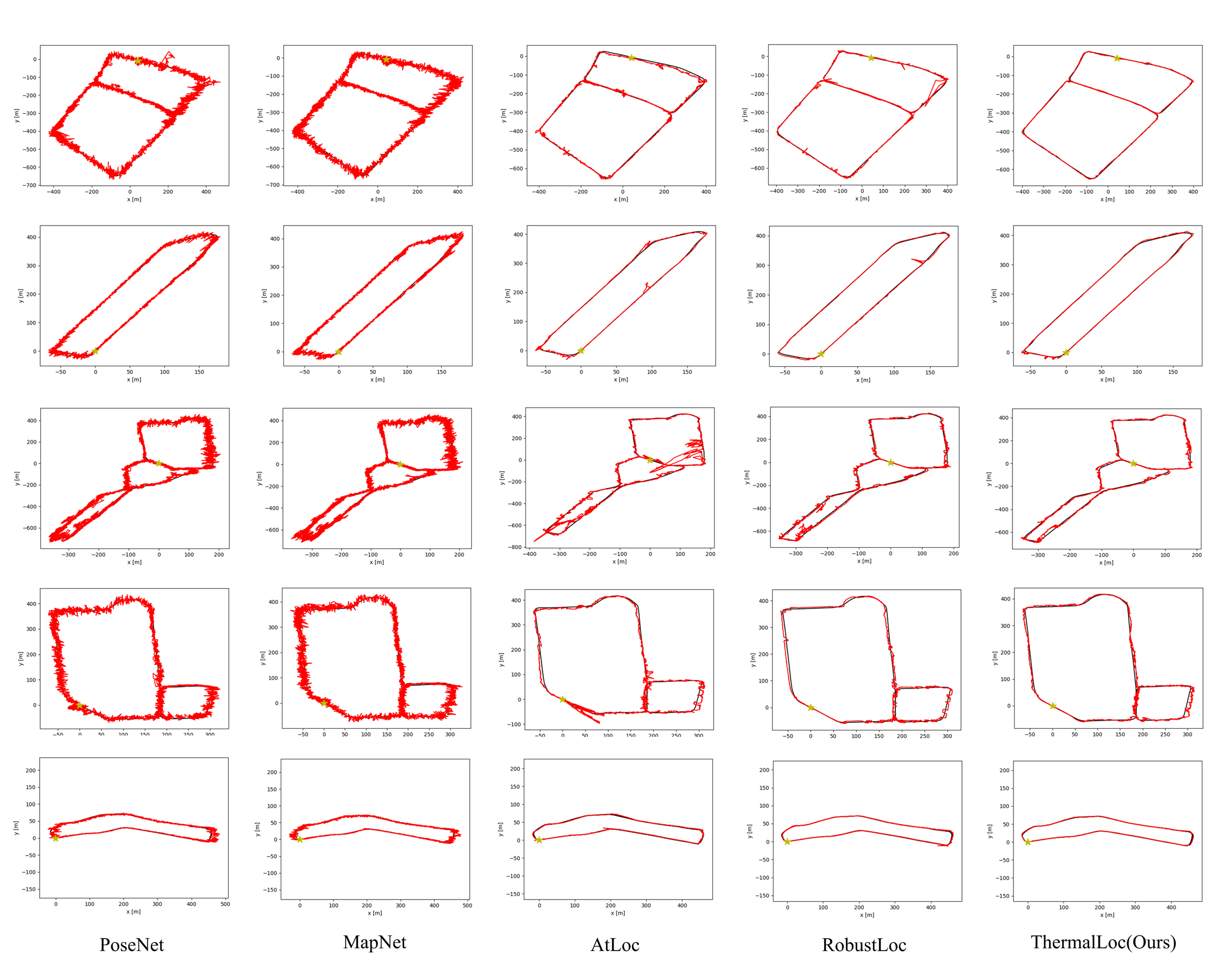}
 \caption{A comparison of the trajectories generated by PoseNet, MapNet, AtLoc, RobustLoc, and ThermalLoc across four scenarios (from top to bottom: IF-1, IF-2, IF-3, IF-4 from our large-scale thermal relocalization dataset and Valley from the public thermal images dataset STheReO \cite{sthereo}). The black line represents the ground truth, the red line denotes the trajectory predicted by each algorithm, and the asterisk marks the start and end points of the trajectory. It is clear that ThermalLoc’s trajectories are consistently closer to the ground truth.}
 \label{Trajectories map}
\end{figure*}

\section{RELATED WORK}
\subsection{Thermal SLAM}
Thermal SLAM remains challenging due to the nature of infrared imaging, which captures heat distribution rather than appearance and geometry. Some approaches combine thermal odometry with other sensors like LiDAR \cite{lidar1, lidar2}, radar \cite{radar}, RGB cameras \cite{visible}, and inertial sensors \cite{inertial1, inertial2}. 
Borges and Vidas developed a practical thermal odometry method by automating the Non-Uniformity Correction (NUC) process, triggered based on attitude predictions \cite{borges}. Chen et al. proposed a thermal-depth odometry method that combines thermal and depth data along with NUC to improve accuracy \cite{chen2023thermal}. Zhao et al. introduced a deep learning-based thermal-inertial odometry approach for thermal feature detection \cite{zhao}, while Saputra et al. developed DeepTIO, a deep learning model for thermal-inertial odometry that addresses drift issues using a loop closure mechanism \cite{deeptio, deeptio2}.
Mouats et al. employed a Fast-Hessian feature extractor for stereo thermal odometry in UAV tracking \cite{inertial2}, and Khattak et al. proposed the Thermal Inertial Odometer, which minimizes radiation errors between keyframes \cite{inertial1, khattak2}. Jiang et al. enhanced the real-time performance of thermal-inertial SLAM with their ThermalRAFT system, which utilizes Singular Value Decomposition (SVD) and a lightweight flow network for improved efficiency \cite{jiang}. Despite these advancements, no effective solution yet exists for large-scale thermal camera relocalization, highlighting an ongoing gap in this field.

\subsection{Deep Learning-Based Camera Relocalization}
Deep learning has greatly advanced camera relocalization by enabling accurate pose predictions in complex environments. Kendall et al. \cite{BYS-based-posenet} pioneered the use of CNNs for end-to-end pose estimation, bypassing traditional feature matching steps. E-PoseNet \cite{eposenet} introduced a lightweight version of PoseNet \cite{posenet}. Approaches such as AtLoc \cite{Atloc}, EffLoc \cite{effloc}, CT-Loc \cite{ctloc}, and TransBoNet \cite{transbonet} improve 2D-to-3D relationships using self-attention mechanisms, while MMLNet \cite{MMLNet} adds reconstruction branches. TransPoseNet \cite{transposenet} combines CNNs with Transformers for depth-based pose estimation, and FeatLoc \cite{featloc} directly employs sparse features for this task.
Other methods, e.g., RadarLoc \cite{radarloc}, SGLoc \cite{sgloc}, NIDALoc \cite{nidaloc}, and STCLoc \cite{stcloc}, leverage different sensor modalities, such as radar and LiDAR, to achieve real-time and precise localization. Zhou et al. introduced DarkLoc, a thermal image-based indoor localization method that uses attention models and relative constraints \cite{darkloc+}. However, most deep learning-based relocalization methods focus on RGB and LiDAR, with the only work (i.e. \cite{darkloc+}), addressing thermal images—and that, too, only in small indoor areas. In contrast, we are the first to explore large-scale thermal camera localization using deep learning on thermal images.

\section{Thermal Camera Relocalization}
As depicted in Figure 2, our proposed ThermalLoc framework is an efficient vision transformer-based method for thermal camera relocalization. Thermal images often face challenges such as low resolution, poor contrast, limited detail, sparse texture information, and edge noise compared to RGB images. To address these issues, we first preprocess the raw thermal images using linear transformation based histogram equalization and a Gaussian low-pass filter. Our ThermalLoc then takes the transformed images as input and leverages an end-to-end combination of EfficientNet and Transformer networks. The EfficientNet module extracts local features, while the Transformer module aggregates and correlates these features, generating multi-scale representations through a self-attention mechanism. Finally, the thermal camera pose estimation module employs two multi-layer perceptrons (MLPs) to regress pose vectors, specifically the 3-dimensional position and the 4-dimensional quaternion-based orientation.

\subsection{Thermal Image Transformation}
Thermal cameras primarily display temperature distribution maps using color coding, where the thermal image corresponds to the temperature of the measured environment or object. While this characteristic facilitates temperature analysis, it can sometimes lead to a loss of scene appearance details. To enhance the feature details of thermal images, we perform several transformations, including image brightness adjustment, contrast adjustment, and detail enhancement. Brightness adjustment ensures that the brightness range is fixed according to the grayscale level, preventing flickering in the collected 10Hz thermal dynamic data. Contrast adjustment primarily stretches the grayscale range, enhancing the temperature differences between various objects in the same image. This process makes edge information more distinct and strengthens texture and detail features. Both brightness and contrast adjustments are achieved through a linear transformation of the image's grayscale values, defined by the following equation: 
define raw thermal image $\mathbf{T}$ 

\begin{equation} 
\mathbf{P^{\prime}}=a\cdot \mathbf{P}+b
\end{equation} 

Here, $\mathbf{P}$ represents the grayscale value of the thermal camera based on temperature imaging $\mathbf{T}$, and $P'$ is the grayscale value after stretching. The parameter $a$ adjusts the contrast, while $b$ controls the brightness.

To further enhance the texture detail information in thermal images, we apply Gaussian low-pass filtering to sharpen thermal image details:
\begin{equation} 
\mathbf{T^{\prime}}=\mathbf{P^{\prime}} +h\times(\mathbf{P^{\prime}} - \mathbf{P^{\prime}} * \mathbf{G})
\end{equation} 

Here, $\mathbf{P^{\prime}}$ represents the image after brightness and contrast adjustments, and $\mathbf{T^{\prime}}$ denotes the final processed result. The parameter $h$ is a manually set sharpening intensity coefficient, while $\mathbf{P^{\prime}} * \mathbf{G}$ represents the convolution of $\mathbf{P^{\prime}}$ with the Gaussian kernel $\mathbf{G}(x,y) = \frac1{2\pi\sigma^2}e^{-\frac{x^2+y^2}{2\sigma^2}}$ at the corresponding coordinate point. The parameter $\sigma$ controls the intensity of the Gaussian filter. As shown in Figure 2 (a), the processing techniques applied to thermal images significantly enhance the representation of detailed texture features.

\subsection{Efficient Transformer Based Thermal Feature Extraction} 
To effectively extract thermal features that capture both local details and global associations, we propose a hybrid feature extractor that combines a CNN with a Transformer. Unlike previous CNN-based camera relocalization models like PoseNet and AtLoc, which excel in extracting detailed features but struggle with global feature associations, Vision Transformers (ViTs) are strong in global feature extraction. However, ViTs typically require more data and computation due to their lack of inductive bias, making them less effective at capturing local information.
Inspired by ViNT \cite{vint} and RT-1 \cite{rt1}, we incorporate a hybrid approach that leverages the strengths of both CNN (EfficientNet) and Transformer architectures, enabling efficient and effective large-scale thermal relocalization.
Our Transformer differs from conventional models in two key ways: (1) we remove the masked attention mechanism normally found in the ViT model, and (2) instead of using multiple linear transformations to compute $\mathbf{Q}(\text{query})$, $\mathbf{K}(\text{key})$, and $\mathbf{V}(\text{value})$, we use a single linear transformation, unlike Transformer \cite{transformer} and RT-1 \cite{rt1}. In our experiments, these two modifications significantly improve pose estimation performance.

\subsubsection{EfficientNet Module}
EfficientNet employs a unique scaling method known as compound scaling, which enhances both efficiency and adaptability. This method is the result of a coordinated neural architecture search (NAS) that simultaneously optimizes for accuracy and efficiency, leading to models that excel across various tasks, particularly in the feature extraction of thermal images. The feature extraction prowess of EfficientNet is attributed to its use of Inverted Residual Blocks, which is shown in Figure 2 (c). The MBConv blocks integrate Depth-wise Convolution (DWConv) and Squeeze-and-Excitation (SE) modules to minimize parameters while improving channel representation. The main difference between MBConv1 and MBConv6 lies in the values of s1 and k in the architecture of MBConv block in Figure 2 (c). The value of s1 is 1 and k is 3 in MBConv1 block, while s1 is 6 and k is 3 or 5 in MBConv6 block, and the value of s2 is usually 4 in both blocks. The MBConv block is as follows:
\begin{equation} 
\mathbf{y}=\mathbf{x}+\text{Dropout}(\text{Conv}(\text{SE}(\text{DWConv}(\text{Conv}(\mathbf{x})))))
\end{equation} 

Here, $\mathbf{y}$ is the feature map from an MBConv block, which feeds into the Transformer module after multiple MBConv layers. The input $\mathbf{x}$ of the first MBConv block is obtained from the processed thermal image $\mathbf{T^{\prime}}$. 

\subsubsection{Transformer Module}
 The Transformer module excels at capturing long-range dependencies in images through its self-attention mechanism, making it particularly effective for global feature association. This capability is especially beneficial for camera relocalization in large-scale scenes. The Transformer block introduced in our approach comprises an Attention module and a FeedForward module, as illustrated in Figure 2 (d). Unlike the standard Transformer, which generates Q, K, and V from separate layers, our method derives them from a single layer and subsequently rearranges them. The self-attention mechanism is as follows:
\begin{equation} 
\mathrm{attn}(\mathbf{y})=\mathrm{softmax}((\mathbf{Q} \cdot \mathbf{K}^T)\times \mathrm{scale})\cdot \mathbf{V}
\end{equation}

Where $\mathbf{y}$ is obtained by the EfficientNet feature map through dimensional transformation and block embedding operations. The FeedForward module is a standard neural network with two linear layers and a GELU activation function, given by:
\begin{equation} 
\mathrm{f}_\text{forward}(\mathbf{o})=\mathrm{Linear}(\mathrm{Dropout}(\mathrm{GELU}(\mathrm{Linear}(\mathbf{o}))))
\end{equation}

Each Transformer module connects the Attention and FeedForward modules with residual connections. The final feature map $\mathbf{f}$, after processing through six Transformer blocks, is:
\begin{equation} 
\mathbf{z}=\mathrm{f}_\text{forward}(\mathrm{attn}(\mathbf{y})+\mathbf{y})+\mathrm{attn}(\mathbf{y})+\mathbf{y}
\end{equation} 

Where $\mathbf{z}$ represents the output of a transformer module, and the output after 6 blocks is represented as $\mathbf{z}^{6}$. The feature map $\mathbf{F}$, which is obtained by the entire transformer feature extraction module is expressed as follows:
\begin{equation} 
\mathbf{F}=\mathrm{Transformer(\mathbf{y})}=\mathrm{Norm(\mathbf{z}^{6})}
\end{equation} 

This hybrid CNN-Transformer approach captures both local and global features, improving relocalization accuracy and enhancing model generalization, with flexibility and scalability for various tasks and datasets.

\subsection{Learning Thermal Camera Pose}
After extracting thermal features using EfficientNet and Transformer, the 6-DoF camera pose is predicted via an MLP-based regressor, given an input thermal image $\mathbf{T^{\prime}}$:
\begin{equation} 
[\mathbf{l},\mathbf{q}]=\text{MLPs}(\text{Transformer}(\text{EfficientNet}(\mathbf{T^{\prime}})))
\end{equation} 

Here, $[\mathbf{l}, \mathbf{q}]$ represents the predicted position $\mathbf{l} \in \mathbb{R}^{3}$ and rotation quaternion $\mathbf{q} \in \mathbb{R}^{4}$ after a forward pass through the network. The model is optimized by minimizing the L1 loss between the predicted values and the ground truth labels $\hat{\mathbf{l}}$ and $\hat{\mathbf{q}}$, as defined by the following loss function \cite{posemlp}:
\begin{equation} 
loss(\mathbf{T^{\prime}})=||\mathbf{l}-\hat{\mathbf{l}}||_{1}e^{-\beta}+\beta+\|\log\mathbf{q}-\log\hat{\mathbf{q}}\|_{1}e^{-\gamma}+\gamma
\end{equation} 

Here, $\beta$ and $\gamma$ are learnable parameters that balance the contributions of position and rotation during training. The logarithm of the unit quaternion $\log\mathbf{q}$ is used instead of the quaternion $\mathbf{q}$ directly, to avoid normalization issues in the L1 loss function. This approach reduces the impact of outliers, enhances robustness to atypical observations, and encourages parameter and feature sparsity. Specifically, the unit quaternion $\mathbf{q}$ consists of a real part $u$ and an imaginary part $\mathbf{v}$. The logarithmic transformation is defined as:
\begin{equation} 
\log\mathbf{q}=\begin{cases}\frac{\mathbf{v}}{\|\mathbf{v}\|}\cos^{-1}u,&\text{if}\|\mathbf{v}\|\neq0\\\mathbf{0},&\text{otherwise}\end{cases}
\end{equation} 

Rotation vectors are mapped to 3D space and converted to normalized unit quaternions, which represent rotation for camera pose estimation. However, due to the non-uniqueness of quaternions (i.e., $-\mathbf{q}$ and $\mathbf{q}$ can represent the same rotation), we constrain all quaternions to a single hemisphere to ensure a unique representation.

\section{EXPERIMENTS}
\subsection{Experiment Setup}
\begin{table}[t]
\centering
\caption{Details of our evaluation dataset used for training and testing.}
\begin{tabular}{c|cccc}
\hline
Sequence & Season & Tag & Distance & Total Image Number \\
\hline
\hline
IF-1 & Spring & Daytime & 3.5km & 22329 \\
IF-2 & Summer & Nighttime & 1.0km & 7735 \\
IF-3 & Summer & Nighttime & 3.3km & 20828 \\
IF-4 & Summer & Nighttime & 1.7km & 11521 \\ 
{Valley} & Summer & All-day & 2.08km & 7860 \\
\hline
\end{tabular}
\label{dataset}
\end{table}

\begin{table*}[t]
\centering
\caption{Camera relocalization results for five scenarios on the thermal dataset. The median and mean errors for both position and orientation are computed for each trajectory using PoseNet, AtLoc, RobustLoc and our proposed ThermalLoc.}
\begin{tabular*}{\textwidth}{@{\extracolsep{\fill}}c|cc|cc|cc|cccccccccc} 
\hline
\multicolumn{1}{c|}{Sequence} & \multicolumn{2}{c|}{PoseNet} & \multicolumn{2}{c|}{AtLoc} & \multicolumn{2}{c|}{RobustLoc} & \multicolumn{2}{c}{ThermalLoc} \\ \hline
\hline
- & Median & Mean & Median & Mean & Median & Mean & Median & Mean \\
IF-1 & 11.31m,2.74° & 13.36m,4.82° & 4.77m,1.05° & 7.26m,3.24° & 3.77m,1.16° & 5.34m,2.91° & \textbf{3.17m,0.55°} & \textbf{4.18m,1.93°} \\
IF-2 & 5.15m,13.23° & 5.93m,16.59° & 4.35m,\textbf{10.86°} & 5.71m,15.13° & \textbf{2.69m},11.86° & 3.50m,16.30° & 2.77m,11.91° & \textbf{3.48m,14.81°} \\
IF-3 & 11.04m,6.21° & 13.51m,12.16° & 5.96m,4.52° & 9.24m,10.77° & 6.19m,4.40° & 6.76m,10.33° & \textbf{5.60m,4.24°} & \textbf{6.10m,9.09°} \\
IF-4 & 7.39m,4.73° & 8.54m,13.84° & 7.33m,4.03° & 7.95m,13.11° & 5.39m,\textbf{2.32°} & 5.86m,12.31°& \textbf{4.62m},2.41° & \textbf{5.27m,11.84°} \\ 
{Valley} & 5.16m,2.06° & 6.34m,3.71° & 3.26m,1.42° & 3.69m,2.64° & 3.16m,0.94° & 3.44m,2.74°& \textbf{1.81m,0.72°} & \textbf{2.10m,1.53°} \\ \hline
Average & 8.01m,5.79° & 9.54m,10.22° & 5.13m,4.38° & 6.77m,8.98° & 4.24m,4.14° & 4.98m,8.92° & \textbf{3.59m,3.97°} & \textbf{4.23m,7.84°}\\ 
\hline
\end{tabular*}
\end{table*}

\begin{table*}[t]
\centering
\caption{Comparison of mean errors of position and rotation among various depth Transformer modules and the Masked Transformer Module in the IF-1 scenario.}
\begin{tabular*}{\textwidth}{@{\extracolsep{\fill}}c|ccccccccc}
\hline
Model & Depth=1 & Depth=2 & Depth=3 & Depth=4 & Depth=5 & Depth=6 (ThermalLoc) & Mask-based \\
\hline \hline
IF-1 & 13.56m,3.69° & 8.55m,2.69° & 5.61m,2.34° & 7.13m,2.47° & 4.65m,2.06° & \textbf{4.18m,1.93°} & 16.35m,3.78° \\
\hline
\end{tabular*}
\end{table*}

\subsubsection{Implementation Details}
In this experiment, each scenario consists of three trajectories collected at different times, with two used for training and the remaining one for testing. The EfficientNet-B0 component of our network is initialized with a pretrained model from the ImageNet dataset, while the other components are initialized randomly. The features extracted by EfficientNet are then passed into the Transformer module for further processing. The network is trained from scratch for 300 epochs using an NVIDIA RTX 3090 GPU.
During testing, each image is processed in 6 milliseconds on an NVIDIA RTX 4060 GPU. With an infrared camera operating at a frequency of 10 Hz, our ThermalLoc model enables real-time predictions. We utilize the ADAM optimizer with a learning rate of $5 \times 10^{-5}$, a batch size of 8, a dropout rate of 0.1, and weight initializations of $\beta_{0} = -3.0$ and $\gamma_{0} = 0.0$.

\subsubsection{Evaluation Datasets}
We evaluated ThermalLoc using our self-collected dataset and the publicly available STheReO \cite{sthereo} dataset.
Our dataset, collected across city-scale environments from four distinct scenes, includes diverse conditions such as different seasons (spring and summer), lighting (day and night), and traffic dynamics. The dataset contains thermal images with a resolution of 480 × 270 pixels, with ground truth position and orientation obtained via high-precision RTK differential positioning (accuracy: 0.1 meters). It also features dynamic elements like vehicles, cyclists, and pedestrians, adding challenges to the relocalization task.
We also used the STheReO dataset, selecting the Valley sequences with a resolution of 640 × 512 pixels. The Valley morning and evening sequences were used for training, and the Valley afternoon sequence for testing.
An overview of the evaluation datasets is provided in Table \textbf{\uppercase\expandafter{\romannumeral1}}.

\subsection{Large-Scale Thermal Camera Relocalization in Urban Environments}
To evaluate our method, we compared ThermalLoc with established end-to-end learning-based visual localization approaches, including PoseNet (CNN based), MapNet (CNN based),  AtLoc (CNN+Attention based) and RobustLoc (CNN+Transformer based).
These end-to-end methods can be directly applied to processed thermal images without considering the transformation of image modes. They address the challenges posed by the lack of texture and detail in thermal images. Unfortunately, since DarkLoc is not open-source, we were unable to include it in our comparison.

To ensure a fair evaluation, we used the same dataset for training and testing across all algorithms, employing position error and rotation error as the evaluation metrics. Table \textbf{\uppercase\expandafter{\romannumeral2}} presents a comparison of our proposed method with PoseNet, AtLoc and RobustLoc on our thermal dataset and the STheReO dataset.
Our results demonstrate that ThermalLoc significantly improves positioning accuracy compared to PoseNet. For instance, in the IF-1 scenario, the mean position accuracy improved from 13.36 meters to 4.18 meters. In the Valley sequences, the mean position accuracy improved from 6.34 meters to 2.10 meters, corresponding to improvements of 68.7\% and 66.9\%, respectively. For a more intuitive comparison, we selected the average performance across five scenarios. Overall, ThermalLoc showed substantial improvements in positioning accuracy over other models. Compared to PoseNet, ThermalLoc reduced the mean position error by 55.7\%. Compared to AtLoc, the reduction was 37.5\%, and compared to RobustLoc, it was 15.1\%.

Figure 3 shows the positioning trajectories of various methods across four scenarios (from top to bottom: IF-1, IF-2, IF-3, IF-4, and Valley). The results clearly indicate that ThermalLoc demonstrates superior smoothness and stability, closely following the ground truth.

\begin{table}[t]
\centering
\caption{Comparison of Mean Errors of position and rotation among ResNet-based, EfficientNet-only and AtLoc computed in the IF-1 scenario}
\begin{tabular}{c|cccc}
\hline
Model & ResNet-based & EfficientNet-only & AtLoc\\
\hline \hline
IF-1 & 7.51m,\textbf{2.26°} & \textbf{5.01m},2.53° & 7.26m,3.24° \\
\hline
\end{tabular}
\end{table}

\subsection{Ablation Study}
In this part, we conducted an ablation study to evaluate the impact of different architectural components on the ThermalLoc model primarily using the IF-1 scenario from our thermal dataset.

\subsubsection{Ablation of the EfficientNet Module}
To assess the contribution of the EfficientNet module in our ThermalLoc model, we performed an ablation study. As shown in Table \textbf{\uppercase\expandafter{\romannumeral4}}, the term "ResNet-based" refers to replacing the EfficientNet module in ThermalLoc with a ResNet module, while keeping all other components of ThermalLoc unchanged. "EfficientNet-only" denotes using only the EfficientNet module as the backbone for feature extraction from thermal images.
The results indicate that replacing EfficientNet with ResNet in ThermalLoc significantly reduces performance. Specifically, ThermalLoc achieves 44.3\% better positioning accuracy than the ResNet-based model in the IF-1 scenario. Furthermore, the EfficientNet-only configuration outperforms AtLoc in relocalization accuracy, demonstrating the suitability of EfficientNet for feature extraction from thermal images.

\begin{table}[t]
\centering
\caption{Comparison of Mean Errors of position and rotation among Patch-First, Shape-First and Vint-Like on our thermal dataset}
\begin{tabular}{c|ccc}
\hline
Model & Patch-First & Shape-First (ThermalLoc) & Vint-Like \\
\hline \hline
IF-1 & 6.53m,2.06° & \textbf{4.18m,1.93°} & 17.30m,3.77° \\
IF-2 & 3.92m,16.48° & \textbf{3.48m,14.81°} & 10.76m,18.07° \\
IF-3 & 7.95m,9.83° & \textbf{6.10m,9.09°} & 19.41m,12.51° \\
IF-4 & 6.05m,12.11° & \textbf{5.27m,11.84°} & 11.45m,16.29° \\
\hline
Average & 6.11m,10.12° & \textbf{4.76m,9.42°} & 14.73m,12.66°\\
\hline
\end{tabular}
\end{table}

\subsubsection{Ablation into Bridging Module}
To improve the integration of EfficientNet and Transformer, we conducted an ablation study on the bridging module. Table \textbf{\uppercase\expandafter{\romannumeral5}} shows the effect of different Bridging modules on our thermal dataset.
\begin{itemize}
    \item Patch-First: This method processes the EfficientNet feature map into patches, converts them into embeddings, and then adjusts the embedding dimensions to fit the Transformer model.
    \item VINT-Like \cite{vint}: This method follows the connection strategy outlined in VINT, where the feature map is masked before being input into the Transformer model.
    \item Shape-First (ThermalLoc): This approach first rearranges the feature map to match the dimensions required by the Transformer model, followed by embedding it into the Transformer.
\end{itemize}

The results demonstrate that the Shape-First method in our ThermalLoc model largely outperforms the other two approaches, indicating that our bridging method is effective.

\subsubsection{Ablation into Transformer Module}
Since the primary goal of our backbone is to extract features from thermal images, the Transformer module was primarily adapted from the Vision Transformer (ViT). To determine the most effective Transformer configuration, we performed a series of ablation studies on different Transformer types and models with varying depths.
As shown in Table \textbf{\uppercase\expandafter{\romannumeral3}}, "Mask-based" refers to using MaskedGoalAttention to mask the feature map. "Depth = n" indicates the use of an n-layer Transformer to extract global feature associations from the image.
The results indicate that the Transformer module without masking performs better in thermal image relocalization, and a 6-layer Transformer strikes an optimal balance between efficiency and effectiveness.

\section{CONCLUSIONS}
This work presents ThermalLoc, which leverages EfficientNet and Transformer to effectively tackle the challenges of thermal camera relocalization in large-scale urban environments. By utilizing an end-to-end deep neural network, ThermalLoc enables robust feature extraction from thermal images, even with minimal texture and detail. The model’s integration of multi-scale features ensures resilience to variations in illumination and dynamic conditions. Our results show that ThermalLoc achieves state-of-the-art performance in monocular thermal camera relocalization. In the future, we aim to extend our model to handle longer time spans and greater distances.







\bibliography{refence}
\bibliographystyle{IEEEtran}

\end{document}